\documentclass[10pt,twocolumn,letterpaper]{article}

\usepackage{cvpr}              

\usepackage[most]{tcolorbox} 
\usepackage{fancyvrb}
\usepackage{fvextra}
\usepackage{array} 
\usepackage{multirow}
\usepackage{booktabs}
\usepackage{caption}
\usepackage[table]{xcolor}
\usepackage{subcaption} 

\definecolor{cvprblue}{rgb}{0.21,0.49,0.74}
\usepackage[pagebackref,breaklinks,colorlinks,allcolors=cvprblue]{hyperref}

\def\paperID{10478} 
\def\confName{CVPR}
\def\confYear{2026}

\title{Chain-of-Adaptation: Surgical Vision-Language Adaptation with \\ Reinforcement Learning}

\author{
Jiajie Li \quad
Chenhui Xu \quad
Meihuan Liu \quad
Jinjun Xiong \\
University at Buffalo \\
Buffalo, NY, USA  
}

\begin{document}
\maketitle

\begin{abstract}

Conventional fine-tuning on domain-specific datasets can inadvertently alter a model’s pretrained multimodal priors, leading to reduced generalization. To address this, we propose Chain-of-Adaptation (CoA), an adaptation framework designed to integrate domain knowledge while maintaining the model’s inherent reasoning and perceptual capabilities. CoA introduces a structured reasoning format that enhances domain alignment without sacrificing general multimodal competence by reinforcement learning. Experiments on standard surgical benchmarks, under both in-distribution and out-of-distribution settings, demonstrate that CoA achieves higher accuracy, stronger generalization, and more stable behavior than supervised fine-tuning. Furthermore, ablation studies confirm that CoA effectively preserves the model’s core visual–language abilities, providing a reliable pathway for domain specialization in VLMs.

\end{abstract}    
\section{Introduction}
\begin{figure*}[t!]
    \centering
        \includegraphics[width=\linewidth]{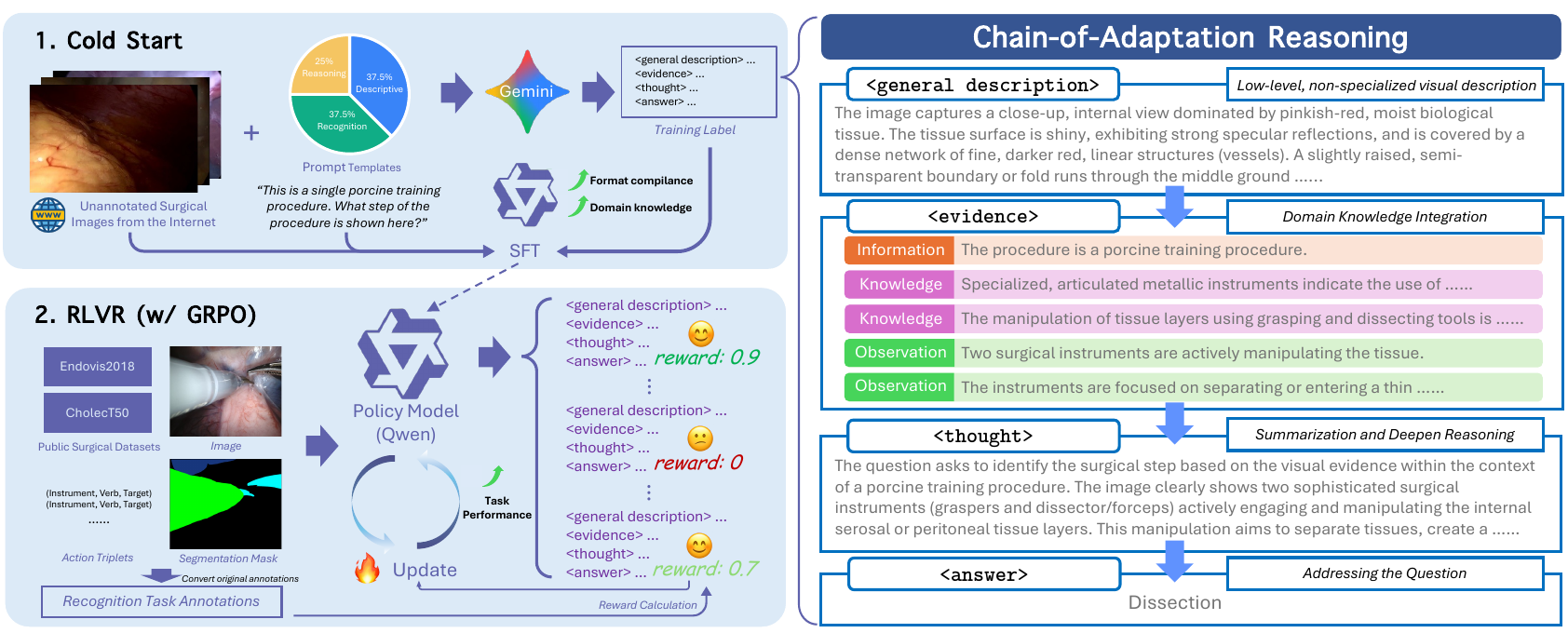}
\captionof{figure}{Chain-of-Adaptation Overview. \textbf{Left:} The training pipeline of the CoA. It features a evidence-orientated cold start that enrich model's domain-specific concepts, and a RL-based training that encourages compliance with the CoA reasoning format and accurate final answers. \textbf{Right:} CoA performs adaptation from general to specialized domains through a four-stage reasoning format.\texttt{\textless general  description\textgreater} gives plain descriptions that language models excel at.\texttt{\textless evidence\textgreater}  explicitly collects mined information and connects it with domain knowledge.\texttt{\textless thought\textgreater}  and \texttt{\textless answer\textgreater} further deepen the reasoning and draw the final conclusion. \vspace{1em}}
        \vspace{-1.5em}
    \label{fig:overview}
\end{figure*}
\label{sec:intro}
Recent advances in vision–language models~(VLMs)~\cite{wang2025internVL35,liu2023visualInstr, Qwen-VL}, have driven substantial progress in visual understanding and multimodal reasoning, enabling models to interpret complex scenes and generate coherent natural-language responses. While in surgical applications, their performance remains limited due to significant distribution shift from general to specific domain. These shifts include (1) the imagery modality and semantic shifts, (2) terminology shifts (e.g. anatomical structures, procedures, instruments), and (3) task requirement shifts (domain-expert-level reasoning).


Towards adapting VLMs to surgical applications, pioneer works including EndoChat~\cite{wang2026endochat}, LLaVA-Surg~\cite{li2024llavasurgmultimodalsurgicalassistant}, GP-VLS~\cite{schmidgall2024gpvls}, and SurgVLM~\cite{zeng2025surgvlm}, have primarily conducted fine-tuning on synthetic instruction-following data that are derived from public surgical datasets' annotations.
Unfortunately, these methods share a common limitation of annotations, i.e., publicly available surgical datasets provide mostly low-level annotations (e.g., categorical labels such as surgical phase, instrument type, or coarse action) while lacking descriptive or reasoning-rich annotations and showing limited linguistic diversity. 
As a results, models are still struggling to generate meaningful responses at scale, either because short-phrase or single word responses lacking semantic richness like in Surgical-VQLA~\cite{bai2023surgical}, or contextually losing like in some LLM-enhanced models~\cite{yang2024llavaMed,li2024llavasurgmultimodalsurgicalassistant}. 
To make matters worse, supervised fine-tuning~(SFT) often generalizes poorly~\citep{zhang2025instructiontuninglargelanguage_sft,pmlr-v267-chu25c}, and frequently causes overfitting to narrow instruction templates and may induce \emph{catastrophic forgetting}~\citep{Kirkpatrick_2017_catastrophic,luo2025empiricalstudycatastrophicforgetting}, or even \emph{model collapse}~\citep{shumailov2024curserecursiontraininggenerated_collapse}, degrading the model’s base reasoning and language abilities.

Fortunately, in practice, current VLMs such as Qwen-VL~\citep{Qwen-VL} already exhibit strong visual grounding, and they can recognize positional, geometric, and temporal attributes of surgical scenes. Leveraging these existing abilities, the core challenge for adapting the VLMs to surgical domains lies not in visual perception, but in the gain of ability to map these observations to clinically meaningful surgical concepts and terminology.
This insight raises a key question: 


\textit{Can we adapt pretrained VLMs to express visual content in clinically accurate ways while leveraging their existing multimodal competence, rather than overwriting it?}

Recent progress in Chain-of-Thought (CoT) Generation~\cite{wei2023chainofthoughtpromptingelicitsreasoning} and reinforcement learning (RL)-based LLM/VLM training~\citep{deepseekai2025deepseekr1incentivizingreasoningcapability,wang2024rlvlmfreinforcementlearningvision_rl_vlm} have demonstrated their  potential to build LLM/VLM's thinking pathways. 
Intuitively, we can construct an adaptation method that transforms from fundamental visual and semantic elements through stepwise reasoning to domain-specific outputs. In terms of form, the model transforms its own extractable low-level visual and linguistic information into domain-specific terminology and concepts through the infusion of specialized knowledge and its own common sense, thereby progressively constructing high-level professional cognition. Specifically, RL methods can fine-tune the model to construct thinking processes and lead to stronger performance by tapping into the model's inherent potential~\cite{yue2025does}, with minimal or no loss of the model's original capabilities. Therefore, in terms of quality, we can adapt the model with RL training to ensure accurate transforms throughout the adaptation process.


As an embodiment of this philosophy, we propose \emph{Chain-of-Adaptation (CoA)}, a RL-based VLM adaptation framework that consists of two components: (1) a structured four-stage reasoning format and (2) a two-stage training pipeline.
The four-stage reasoning format progressively specializes the model’s inference process: it begins with a preliminary description grounded purely in native vision–language abilities,
then refines its observation into precise surgical terminology and integrates domain knowledge to produce clinically grounded evidence,
and finally synthesizes these cues through structured reasoning to deliver an expert-level answer. The training pipeline includes a cold start phase that enriches the model’s surgical knowledge and instills the CoA reasoning format, followed by an RL phase that enhances task performance while preserving its inherent vision–language capability.
Empirically, CoA improves the model's clinical accuracy while preserving its inherent general vision–language capability.
Our experiments show that, on two standard surgical benchmarks, CholecT50~\citep{nwoye2023cholectriplet2022_cholect50} and EndoVis2018~\citep{allan20202018roboticscenesegmentation_endovis2018}, CoA outperforms SFT, improving F$_1$-score from 0.587 to 0.644 (\textbf{+10\%}) and from 0.657 to 0.837 (\textbf{+27\%}), respectively.

\noindent Our main contributions are as follows:
\begin{itemize}
    \item We empirically analyze the limitations of SFT for surgical adaptation, revealing its risk to model collapse, overfitting, and degradation of pretrained linguistic priors.
    \item We introduce \emph{CoA}, a RL–based training framework that aligns VLMs with surgical concepts while maintaining their general multimodal capacity.
    \item We validate our method on multiple standard surgical benchmarks and general-purpose VQA benchmarks, and demonstrate consistent improvements in generalization and robustness over SFT and vanilla CoT-based variant.
\end{itemize}

\section{Related Work}
\begin{figure*}[!t]
  \centering
  \begin{minipage}[t]{0.48\linewidth}
    \vspace{0pt}
    \centering
    \includegraphics[width=\linewidth]{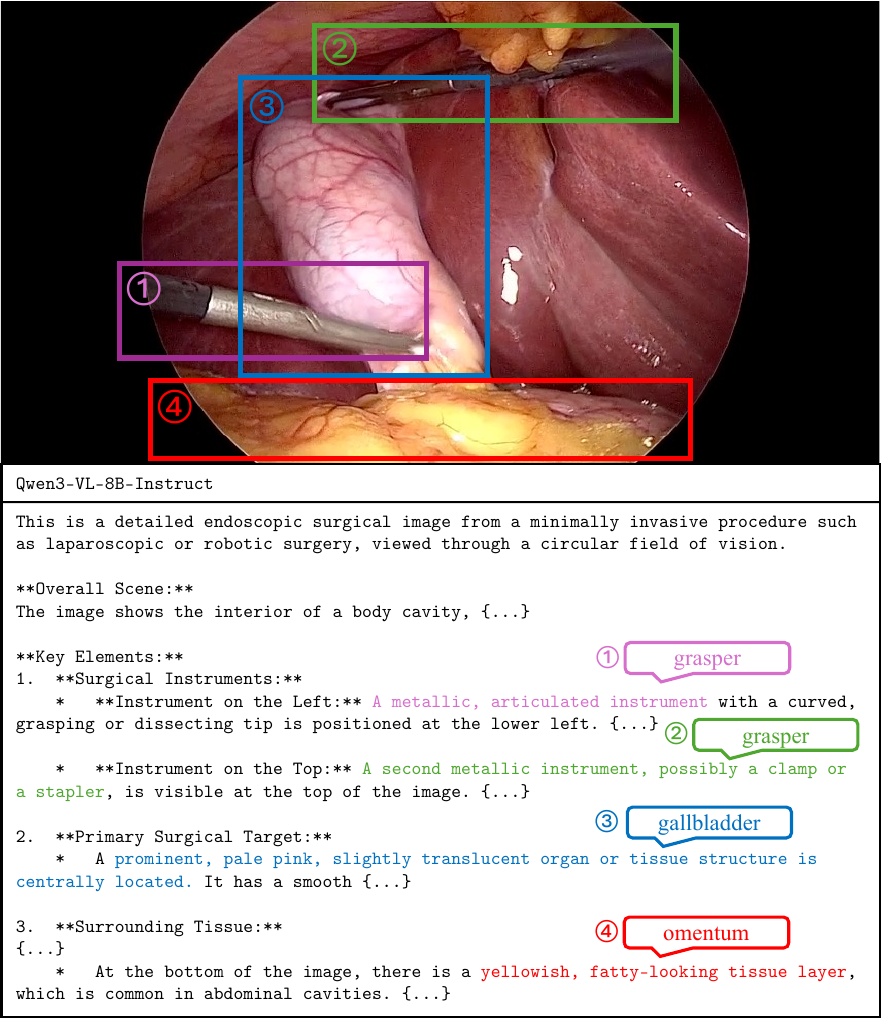}
    \caption{\texttt{Qwen3-VL-8B-Instruct}'s response to: ``Describe the surgical image in detail.''}
    \label{fig:qwen_example}
  \end{minipage}\hfill
  \begin{minipage}[t]{0.48\linewidth}
    \vspace{0pt}
    \centering
    \includegraphics[width=\linewidth]{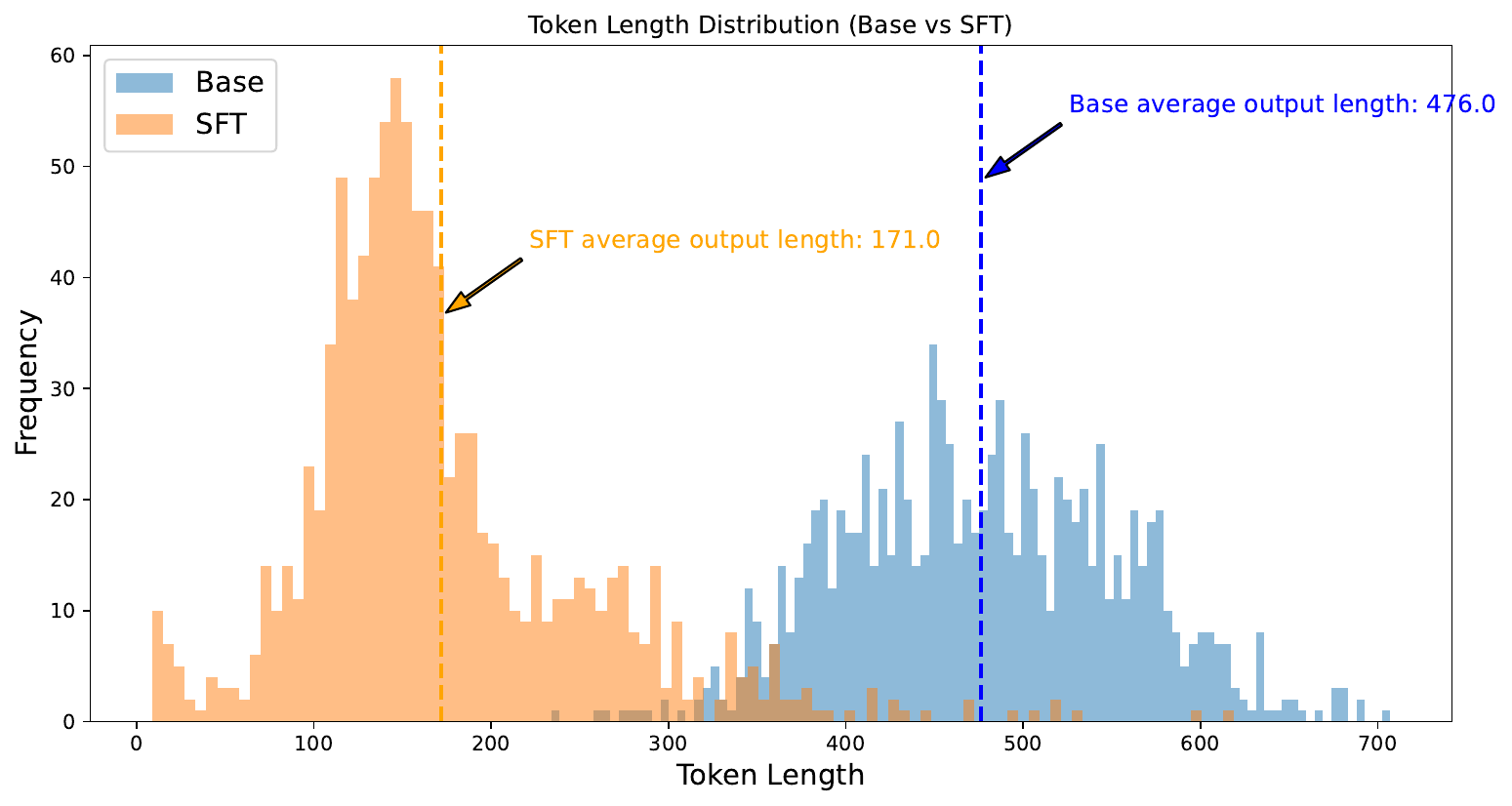}
    \caption{Token Length reduces dramatically after small-scale SFT.}
    \label{fig:token_len}

    \vspace{6pt}
    \includegraphics[width=\linewidth]{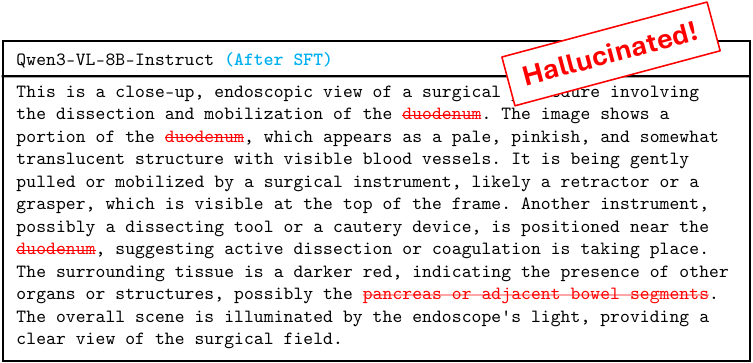}
    \caption{Model's response after SFT.}
    \label{fig:after_sft}
  \end{minipage}
  \vspace{-0.5em}
\end{figure*}

\paragraph{General-Purpose Vision-Language Models.} The widespread adoption of VLMs was accelerated by LLaVA~\citep{liu2023llava}, which popularized the visual instruction tuning paradigm and transformed them into conversational agents. Subsequent models have continuously advanced in scale, data efficiency, and general capability: for example, Qwen-VL~\citep{Qwen2-VL,Qwen-VL} has gained attention for its versatility in fine-grained understanding and localization across various tasks; InternVL~\citep{wang2025internVL35, chen2024internvl} demonstrates the potential for achieving state-of-the-art by scaling up vision encoders and refining the alignment process; meanwhile, closed-source models like Kimi-VL~\citep{kimiteam2025kimivltechnicalreport}, Gemini~\citep{comanici2025gemini25pushingfrontier}, and GPT-4V~\citep{wu2024gpt4visionhumanalignedevaluatortextto3d} leverage their vast scale and massive datasets to showcase powerful general visual abilities and continually set the performance ceiling. However, their pretrained capabilities remain limited in specialized domains like surgery, which require domain-specific semantics and procedural understanding.

\vspace{-0.4cm}

\paragraph{Vision-Language Models with Reinforcement Learning.} Recent work has begun applying the paradigm of Reinforcement Learning from Human Feedback (RLHF)~\citep{ouyang2022traininglanguagemodelsfollow_rlhf} to VLMs. For example, RL-VLM-F~\citep{wang2024rlvlmfreinforcementlearningvision_rl_vlm} uses a VLM to judge image-text pairs and automatically generate reward functions; \citet{chen2024visionlanguagemodelsprovidepromptable} fine-tune VLMs as decision-making agents via RL using chain-of-thought (CoT)~\citep{wei2023chainofthoughtpromptingelicitsreasoning} reasoning; G1~\citep{chen2025g1bootstrappingperceptionreasoning} trains VLMs in interactive RL environments to bootstrap perception and reasoning. However, these methods primarily focus on general visual tasks.

\vspace{-0.4cm}

\paragraph{Surgical Vision-Language Models.}
To adapt general VLMs to complex clinical and surgical environments, most prior work relies on SFT with domain-specific instruction data.  
LLaVA-Med~\citep{yang2024llavaMed}, extending LLaVA, uses GPT-4 to synthesize large-scale medical instruction–response pairs for conversational adaptation.   Subsequent efforts such as LLaVA-Surg~\citep{li2024llavasurgmultimodalsurgicalassistant}, GP-VLS~\citep{schmidgall2024gpvls}, and Surgical-LLaVA~\citep{jin2024surgicalLLAVA} using SFT on synthetic QA pairs from surgical datasets, but suffer from limited annotation diversity and overfitting to frequent categories.  
Recent models like SurgVLP~\citep{zeng2025surgvlm} and Surgical-VQLA~\citep{bai2023surgical} employ contrastive or supervised objectives on surgical lecture videos, enabling zero-shot recognition and captioning.  
Recently, Surgery-R1 \citep{hao2025surgery_r1} explores RL for surgical reasoning VLMs, aiming to achieve interpretable and spatially grounded behavior through rule-based rewards. In contrast, our work provides a systematic examination of both SFT- and RL-based adaptation strategies in surgical domains, beyond performance metrics. We emphasize on preserving the general visual–language competence of VLMs during surgical adaptation.

\vspace{-0.4cm}

\paragraph{Existing Surgical Datasets.} 
Publicly available surgical datasets are predominantly designed for low-level visual understanding and can be broadly categorized into two types: 
(1) Recognition tasks~\citep{twinanda2016endonetdeeparchitecturerecognition, stauder2016tum, maier2021heidelberg, huaulme2021micro, wang2022autolaparo, wagner2023comparative, nwoye2023cholectriplet2022_cholect50, ayobi2024pixelwiserecognitionholisticsurgical_grasp}, 
including phase, step, instrument, action, and organ recognition; and 
(2) Segmentation or detection tasks~\citep{allan20202018roboticscenesegmentation_endovis2018, huaulme2021micro, wang2022autolaparo, ayobi2024pixelwiserecognitionholisticsurgical_grasp, van2022gesture}, 
which provide pixel- or object-level annotations for instruments or anatomical structures. 
While these datasets have facilitated progress in perception-oriented modeling, they largely lack rich linguistic supervision and high-level reasoning signals, 
making them insufficient for aligning VLMs to professional surgical communication.

\section{Dilemma in Surgical VLM Adaptation}

We revisit the challenges of adapting general-purpose vision-language models (VLMs) to surgical scenarios and analyze why this setting poses unique difficulties. Specifically, we examine: (1) how well general-purpose VLMs describe surgical scenes, (2) what is required for effective domain alignment, and (3) whether conventional SFT is sufficient, or even worse, instead introduces new risks.

\subsection{Rethinking SFT in the Surgical Domain}

SFT remains the most common strategy for adapting pretrained models. 
Given a pretrained policy $\pi_\theta(y|x)$, where $x$ denotes multimodal input and $y^*$ as the reference response, SFT minimizes the negative log-likelihood:
\begin{equation}
\mathcal{L}_{\text{SFT}}(\theta) = -\mathbb{E}_{(x, y^*) \sim \mathcal{D}}\!\left[\log \pi_\theta(y^*|x)\right].
\end{equation}
This objective aligns model outputs with labeled references by maximizing the likelihood of human-provided responses. 
While effective in transferring knowledge, this formulation inherently constrains the model to reproduce surface-level linguistic patterns rather than learning broader conceptual alignment. 
As a result, SFT risks narrowing behavioral diversity and weakening previously acquired priors, a risk especially pronounced in the surgical domain, where data are scarce and domain semantics are highly specialized~\citep{luo2025empiricalstudycatastrophicforgetting}.

\subsection{Empirical Observation After SFT}
Modern VLMs, such as \texttt{Qwen3-VL-8B-Instruct}~\citep{qwen3}, already demonstrate strong perceptual grounding in surgical scenes (Fig.~\ref{fig:qwen_example}). They accurately describe geometry, texture, and spatial relations, suggesting that perception is not the bottleneck. The challenge lies in expression to align what the model already ``sees'' with appropriate surgical terms.

To test whether SFT achieves this alignment, we fine-tune \texttt{Qwen3-VL-8B-Instruct} on 700 Q\&A pairs sampled from the \textit{Surg-396K} dataset~\citep{wang2026endochat} for one epoch with a learning rate of $1\times10^{-5}$. We then prompt it to describe 1{,}000 surgical images from CholecT50~\citep{nwoye2023cholectriplet2022_cholect50}. Despite this light setting, substantial behavioral shifts emerge:

\begin{enumerate}
    \item \textbf{Model collapse:} The average output length drops by 64\% (Fig.~\ref{fig:token_len}), accompanied by loss of linguistic diversity, which is consistent with the ``collapse'' phenomenon as reported by~\citet{shumailov2024curserecursiontraininggenerated_collapse}.
    \item \textbf{Overfitting and hallucination:} The model begins to hallucinate domain-specific terms, such as mislabeling non-relevant tissue as \textit{duodenum} (Fig.~\ref{fig:after_sft}).
    \item \textbf{Degraded generalization:} The model’s overall multimodal fluency weakens, producing shorter and less structured descriptions even on general scenes.
\end{enumerate}

These results indicate that even mild SFT can distort the pretrained model’s linguistic space, leading to collapse, hallucination, and reduced general reasoning capacity.

\subsection{Why Does SFT Fall Short?}
Although SFT remains effective in principle, it struggles in data-scarce, expert-heavy domains such as surgery. Existing datasets typically contain only simple phase labels or object-level annotations, while obtaining expert-verified, semantically rich text data is both costly and slow. Consequently, SFT tends to overfit limited supervision rather than fostering genuine conceptual understanding.   This dilemma motivates the need for a more data-efficient and generalizable adaptation framework, one that preserves pretrained multimodal competence while aligning model behavior with domain-specific reasoning.

\section{CoA: Chain-of-Adaptation}

In the previous section, we identified the inherent limitations of SFT under scarce and low-diversity surgical annotations.
In this section, we propose a RL–based framework, termed \textbf{Chain-of-Adaptation (CoA)}, 
which enables the model to efficiently adapt VLMs to surgical tasks by \textit{learning from itself} through iterative answering and feedback. 
Meanwhile, we introduce a novel structured thinking format that maximizes the preservation of the model’s original multimodal competence during adaptation.

\subsection{Preliminary: RLVR and GRPO}

In SFT, the model is trained to directly imitate target responses that demonstrate the desired task behavior by minimizing the negative log-likelihood loss. 
However, this relies on large and diverse annotations, which are often scarce in surgical domains. 
To overcome this limitation, we adopt \textbf{Reinforcement Learning with Verifiable Rewards (RLVR)} as a more scalable adaptation paradigm. 
Instead of mimicking reference outputs, RLVR optimizes the model to favor responses that yield higher, objectively verifiable rewards. 
This formulation is particularly well suited for the surgical domain, where many tasks such as tool recognition, phase classification, or step detection can be automatically assessed using quantitative metrics.
For instance, a response that correctly identifies more instruments or procedural steps receives a higher reward, driving the model toward clinically accurate behavior.

To realize this reward-driven training, we use \textbf{Group Relative Policy Optimization (GRPO)}~\citep{shao2024deepseekmath} as the underlying optimization algorithm. 
Unlike SFT, which relies on absolute supervision, GRPO optimizes for \textit{relative preferences} among multiple self-generated outputs. 
Given an input $x$, the policy $\pi_\theta$ samples a group of responses $\{y_1, y_2, \dots, y_G\}$ from the old policy $\pi_{\theta_{\text{old}}}$, 
each receiving a scalar reward $r_i$ based on its verifiable task performance. 
A normalized advantage $A_i$ is computed as:
\begin{equation}
A_i = \frac{r_i - \text{mean}(\{r_1, r_2, \dots, r_G\})}{\text{std}(\{r_1, r_2, \dots, r_G\})}.
\end{equation}
The policy is updated to favor higher-reward responses while regularizing deviation from the reference distribution:
\begin{equation}
\resizebox{0.90\linewidth}{!}{$
\begin{aligned}
\mathcal{L}_{\text{GRPO}}(\theta)
&= \mathbb{E}_{x} \Bigg[
\frac{1}{G}\sum_{i=1}^{G}
\min \Bigg(
\frac{\pi_\theta(y_i|x)}{\pi_{\theta_{\text{old}}}(y_i|x)} A_i,
\text{clip}\!\left(
\frac{\pi_\theta(y_i|x)}{\pi_{\theta_{\text{old}}}(y_i|x)},
1 - \epsilon, 1 + \epsilon
\right) A_i
\Bigg)
\\[-2pt]
&\quad
- \beta\, D_{\text{KL}}\!\left(\pi_\theta \,\|\, \pi_{\text{ref}}\right)
\Bigg].
\end{aligned}
$}
\end{equation}

Here, the first term reinforces responses with higher relative advantages, 
while the KL regularization prevents over-deviation from the reference model. 
This preference-based alignment allows the model to improve through internal comparison rather than explicit supervision, 
making RLVR particularly effective for surgical adaptation where common task rewards are naturally verifiable.

\subsection{Structured Reasoning: From CoT to CoA}
\label{sec:coa_format}

The conventional chain-of-thought (CoT) paradigm~\citep{deepseekai2025deepseekr1incentivizingreasoningcapability} decomposes reasoning into \texttt{<thought>} ... \texttt{</thought>} and \texttt{<answer>} ... \texttt{</answer>} sections, 
which is effective for logic-intensive tasks such as mathematics or code generation. 
However, for multimodal surgical understanding, the main challenge lies not in reasoning depth but in balancing specialization and generalization under limited supervision. 
To address this, we extend CoT by introducing two additional sections, forming a more structured \textbf{CoA reasoning format}: 

\begin{itemize}
    \item \textbf{\texttt{<general description>}}: This section helps preserve the model’s innate descriptive capability inherited from large-scale pretraining. 
    It functions as a regularization term that maintains general visual–linguistic grounding and mitigates catastrophic forgetting. 
    In this section, the model is explicitly instructed to avoid any surgical or domain-specific terminology, focusing instead on low-level visual attributes such as color, geometry, texture, and spatial relationships.
    \item \textbf{\texttt{<evidence>}}: This section encourages the model to explicitly enumerate reasoning cues by integrating three categories of information: (1) \textit{task information} provided in the prompt, (2) \textit{visual observations} derived from the image, and (3) \textit{domain knowledge} recalled from surgical expertise or medical commonsense. This section not only grounds reasoning in verifiable evidence but also enriches the model’s domain-specific vocabulary and knowledge base during the subsequent cold start phase.
\end{itemize}

\vspace{-1em}
\paragraph{Rationale.}
The right side of \cref{fig:overview} illustrates how the proposed CoA format decomposes surgical reasoning into interpretable components. Structurally separating general visual grounding (\texttt{<general description>}) from domain-specific reasoning (\texttt{<evidence>}) encourages the model to organize information hierarchically rather than mixing perception and reasoning in a single sequence. This explicit decomposition stabilizes optimization by reducing ambiguity in reward attribution and preventing interference between low-level visual semantics and high-level surgical logic. Empirically, as demonstrated in \cref{sec:exp_ablation}, this structured format consistently outperforms vanilla CoT-based variant in both in-distribution and out-of-distribution surgical tasks, as well as general-domain VQA benchmarks.


\subsection{Two-Stage Surgical VLM Adaptation}
To clearly illustrate the overall CoA framework, we provide a high-level overview of the two-stage adaptation pipeline.
The left part of \cref{fig:overview} outlines the complete CoA training process, which consists of two sequential stages: 
(1) a \emph{Cold Start} stage that bootstraps domain knowledge from unlabeled surgical data, and 
(2) a \emph{RLVR} stage that further refines reasoning quality through task-specific feedback. 
Inspired by recent advances in RL-based post-training~\citep{deepseekai2025deepseekr1incentivizingreasoningcapability}, 
this design transforms surgical video adaptation into a structured, verifiable learning process.

\vspace{-.3cm}

\paragraph{Cold Start.} 
We begin with a cold start stage that exposes the model to the CoA reasoning format while enriching its surgical domain knowledge.
As shown in the upper-left of \cref{fig:overview}, we collect about 10K unlabeled surgical images from public internet sources along with their video titles, 
and use a strong multimodal model (\texttt{Gemini-Flash-2.5}~\citep{comanici2025gemini25pushingfrontier}) to automatically generate pseudo-labeled samples following the CoA reasoning format discussed in \cref{sec:coa_format}.
At this stage, correctness is secondary. The goal is to (1) expose the model to structured CoA-style reasoning, and (2) enrich its surgical vocabulary and domain awareness. 
The data cover three task types: surgical scene description, instrument recognition, and reasoning-based analysis (e.g., tool–tissue interaction). 
This process establishes a domain-aware representation prior that facilitates stable RL.

\vspace{-.3cm}

\paragraph{RLVR.}
This stage is the core of the CoA training process, where the model is directly optimized to improve task performance through verifiable feedback. 
Following the GRPO formulation, the model samples multiple candidate outputs for each input and receives scalar rewards reflecting their correctness. 
Each surgical task defines its own reward function $r_{\text{task}}(y) \in [0,1]$, computed from task-specific metrics such as accuracy or F$_1$-score. 
For example, in surgical phase classification or instrument recognition, responses that correctly predict more categories or interactions obtain higher rewards. 
These rewards are normalized into advantages that guide the policy update toward higher-performing responses.

To ensure both structural validity and semantic accuracy, CoA adopts a composite reward function:
\begin{equation}
r(y) =
\begin{cases}
r_{\text{task}}(y), & \text{if } y \text{ conforms to CoA format},\\[4pt]
0, & \text{otherwise.}
\end{cases}
\end{equation}
This formulation enforces both format consistency and task fidelity: malformed outputs receive zero reward, 
while valid ones are ranked by their domain-level correctness. 
During training, the model iteratively generates and evaluates responses, 
and the policy is updated via GRPO to favor high-reward outputs while maintaining stability through KL regularization.

Overall, this stage moves CoA beyond imitation learning, optimizing directly for measurable task success. 
By aligning the objective with verifiable rewards instead of static annotations, 
RLVR enables continuous self-improvement in reasoning and decision quality under limited supervision.

\section{Experiments}
\label{sec:experiments}
\begin{table*}[ht]
\setlength{\tabcolsep}{.5pt} 
\centering
\caption{In-distribution Evaluation on Endovis2018 and CholecT50 (\%)}
\vspace{-.5em}
\label{tab:main}
\begin{tabular}{p{5cm}*{8}{>{\centering\arraybackslash}p{1.5cm}}}
\toprule
 \multirow{2}{*}{Method}  & \multicolumn{4}{c}{Endovis2018} & \multicolumn{4}{c}{CholecT50} \\
\cmidrule(lr){2-5} \cmidrule(lr){6-9}
 & Precision & Recall & F$_1$ & F$_1^{\text{cls}}$ & Precision & Recall & F$_1$  & F$_1^\text{cls}$ \\
\midrule
Base (Qwen3-VL-8B-Instruct) & 54.4 & 45.6 & 48.5 & 42.0 & 26.0 & 65.1 & 35.3 & \textbf{20.7} \\
+ SFT & 65.9 & 69.8 & 65.7 & 43.2 & 60.0 & 62.2 & 58.7 & 15.3 \\
+ Cold Start & 58.0 & 64.2 & 59.3 & 45.5 & 26.8 & 62.1 & 36.2 & 20.4 \\
+ Cold Start + SFT & 63.6 & 64.3 & 62.0 & 45.6 & 64.8 & 65.8 & 62.4 & 15.4 \\
\rowcolor{green!15}
 + Cold Start + RLVR (CoA) & 80.6 & 89.8 & \textbf{83.7} & \textbf{58.0} & 57.6 & 78.4 & \textbf{64.4} & 20.2 \\
\bottomrule
\end{tabular}
\vspace{-.2cm}
\end{table*}



\begin{figure}[t]
  \centering
  \begin{subfigure}{0.48\linewidth}
    \centering
    \includegraphics[width=\linewidth]{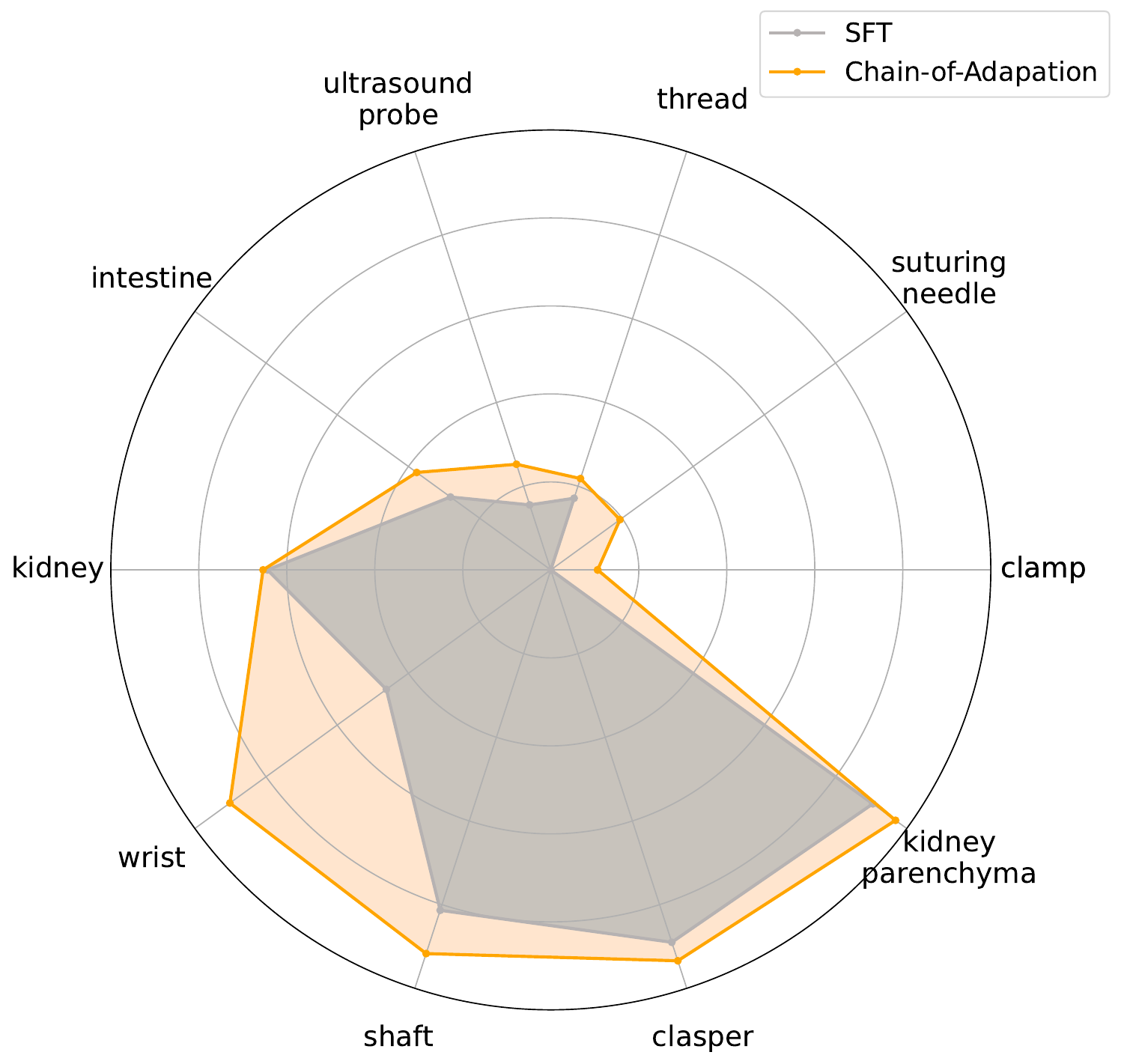}
    \caption{Endovis2018 (F$_{1}$-score)}
    \label{fig:radar_endovis}
  \end{subfigure}
  \hfill
  \begin{subfigure}{0.48\linewidth}
    \centering
    \includegraphics[width=\linewidth]{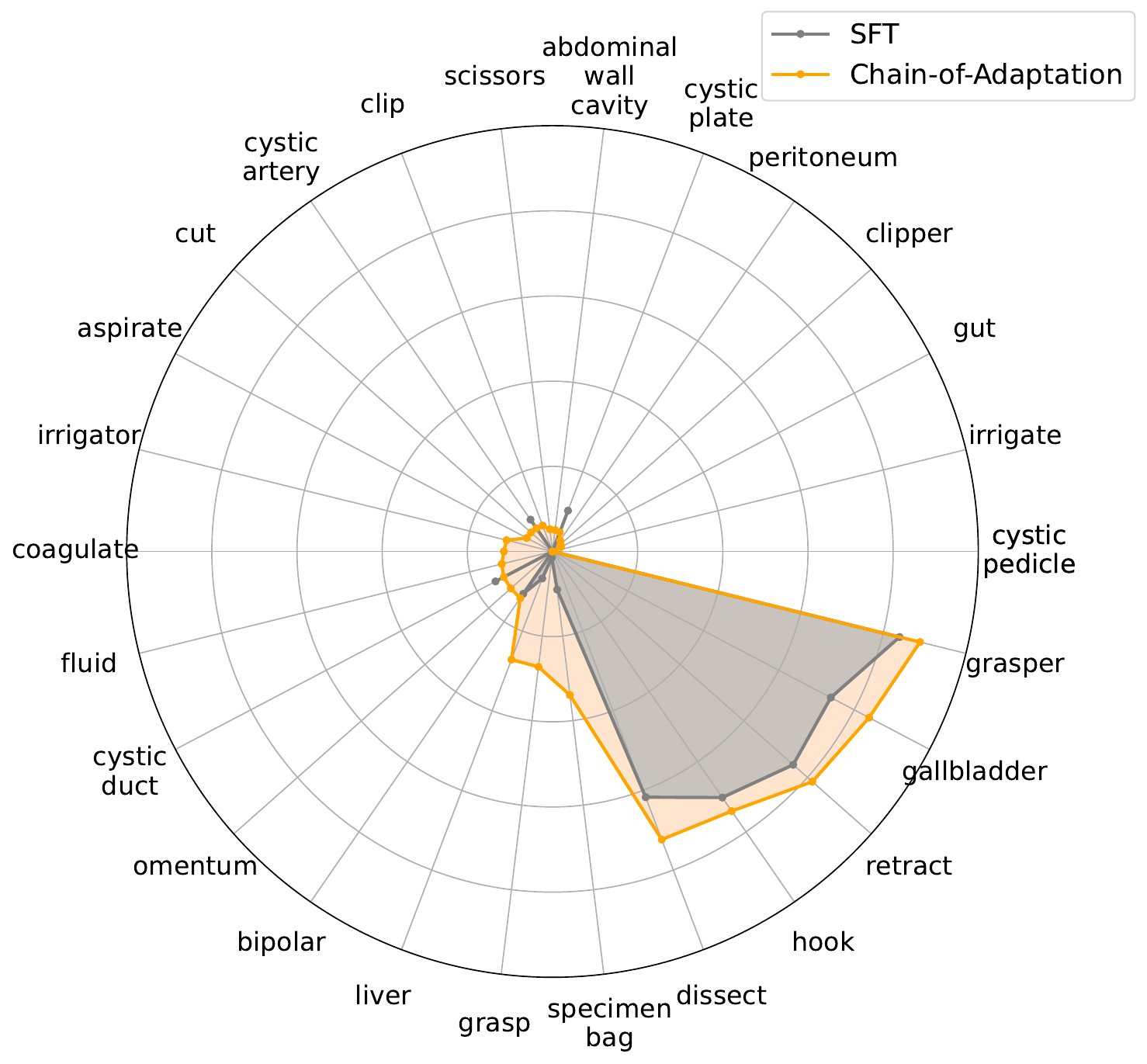}
    \caption{CholecT50 (F$_{1}$-score)}
    \label{fig:radar_cholect50}
  \end{subfigure}

  \caption{SFT vs GRPO on CholecT50 and Endovis2018.}
  \label{fig:radar}
  \vspace{-.5em}
\end{figure}

In this section, we aim to answer two research questions:
(i) Does the proposed CoA framework outperform SFT in surgical vision-language tasks?
(ii) Can CoA’s structured reasoning format enhance performance while mitigating overfitting and preserving general multimodal understanding?
We present the experimental setup and results on various surgical object recognition tasks, followed by evaluations of zero-shot and cross-domain generalization, as well as ablation analyses.

\subsection{Setup and Benchmarks}

\paragraph{Implementation.}
All experiments are implemented using the official QwenVL training framework\footnote{\href{https://github.com/QwenLM/Qwen3-VL/tree/main/qwen-vl-finetune}{https://github.com/QwenLM/Qwen3-VL/tree/main/qwen-vl-finetune}} 
for supervised fine-tuning (SFT) and the SWIFT framework\footnote{\href{https://github.com/modelscope/ms-swift}{https://github.com/modelscope/ms-swift}}~\citep{zhao2024swiftascalablelightweightinfrastructure}
for RLVR optimization. 
Training is performed on 8$\times$NVIDIA H100 GPUs with mixed precision, and all evaluations are run using the vLLM inference engine~\citep{kwon2023efficient_vllm}. 

\vspace{-.35cm}

\paragraph{Datasets.}
We evaluate our method on the task of surgical object recognition using two representative and widely used surgical benchmarks, EndoVis2018~\citep{allan20202018roboticscenesegmentation_endovis2018} and CholecT50~\citep{nwoye2023cholectriplet2022_cholect50} .  
EndoVis2018 is from the MICCAI Endoscopic Vision Challenge, and includes annotated frames of robotic-assisted surgeries with multiple instruments such as graspers, scissors, and clips. There are 2,235 images for training and 996 for testing.
CholecT50 provides diverse laparoscopic cholecystectomy videos with varying viewpoints and instruments. From its training and testing splits, we sample 2,000 frames for training and 1,000 for testing.
For each annotated image, we convert the original recognition annotations into a structured object list and prompt the LLM to identify which entities (i.e., instruments present in the scene) appear in the image. The model selects the relevant entities from a predefined vocabulary (10 classes for Endovis2018 and 28 classes for CholecT50), which we construct by collecting all object categories that appear in the dataset annotations and presenting them to the LLM as a candidate list within the prompt.

\vspace{-.35cm}

\paragraph{Cold start.}
We sample 10{,}000 frames from public surgical lecture videos with
various surgery types. Each image is paired with one of three question types: scene description, object recognition and reasoning (e.g., anomaly detection, and procedural intent). Their sample ratios are 37.5\%, 37.5\% and 25\%, respectively. (Question templates are provided in the supplementary materials.)
Video titles are used as auxiliary context, and \texttt{Gemini-Flash-2.5}~\citep{comanici2025gemini25pushingfrontier} in non-thinking mode generates pseudo-labeled CoA responses.  
We train \texttt{Qwen3-VL-Instruct-8B} on this cold start data for one epoch with a 1e\text{-}5 learning rate to initialize domain awareness before RLVR optimization. Detailed hyperparameters are listed in the supplementary materials.

\vspace{-.35cm}

\paragraph{SFT.} The training for the SFT baseline uses the same hyperparameters as in the cold start. We provide an example of SFT data in the supplementary materials.

\vspace{-.35cm}

\paragraph{RLVR.}
We perform RLVR training for 1 epoch with a learning rate of 1e\text{-}6, total batch size of 112,
and $\beta{=}0.001$ for KL regularization.
Each prompt generates eight responses with a temperature of 1.0.
The optimization follows the GRPO algorithm. Detailed hyperparameters and training dynamics are provided in the supplementary materials.

\vspace{-.35cm}

\paragraph{Metrics.}
We report Precision, Recall, macro-averaged F$_1$, and per-class macro F$_1^{\text{cls}}$ to assess both global and class-balanced performance.  
The class-wise F$_1^{\text{cls}}$ is defined as
\begin{equation}
\text{F}_1^{\text{cls}} = \frac{1}{C} \sum_{c=1}^{C} \frac{2 P_c R_c}{P_c + R_c},
\end{equation}
where $P_c$ and $R_c$ are the precision and recall for class $c$, and $C$ is the total number of classes.  
While F$_1$ reflects overall performance, F$_1^{\text{cls}}$ highlights class imbalance effects. For reasoning models, when computing the metrics (including RLVR and evaluation), we only consider the content within the \texttt{<answer>} section.

\subsection{In-distribution Surgical Evaluation}
\label{sec:exp_sft}
\begin{figure*}[t]
  \centering
  \begin{minipage}[t]{\linewidth}
    \vspace{0pt}
    \centering
    \includegraphics[width=\linewidth]{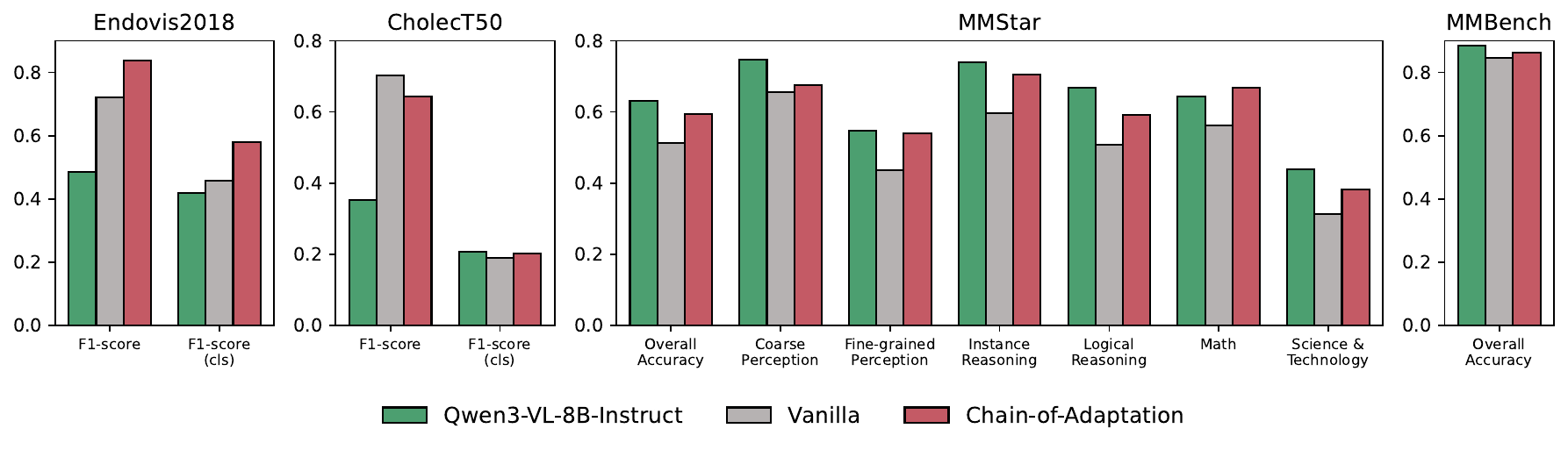}
    \vspace{-.7cm}
    \caption{Performance Comparison on Multiple Datasets}
    \label{fig:ablation_thinking}
  \end{minipage}

  \vspace{12pt}

  \begin{minipage}[t]{0.48\linewidth}
    \centering
    \captionof{table}{Out-of-distribution Evaluation on GraSP (\%)
    }
    \vspace{-.3cm}
\setlength{\tabcolsep}{9pt} 
\centering
\label{tab:ood}
\begin{tabular}{lcccc}
\toprule
\multirow{2}{*}{Method}  & \multicolumn{4}{c}{GraSP}  \\
\cmidrule(lr){2-5}
 & Precision & Recall & F$_1$ & F$_1^{\text{cls}}$ \\
\midrule
Base & 10.8 & 42.9 & 16.4 & \textbf{13.5} \\
SFT & 11.0 & 23.5 & 13.4 & 9.6 \\
\rowcolor{green!15}
 CoA (Ours) & 14.6 & 30.5 & \textbf{18.3} & 13.3 \\
\bottomrule
\end{tabular}

  \end{minipage}\hfill
  \begin{minipage}[t]{0.48\linewidth}
    \centering
    \captionof{table}{SFT vs RLVR (w/o thinking) (\%)}
    \vspace{-.3cm}
\setlength{\tabcolsep}{8pt} 
\centering
\label{tab:nothinking}
\begin{tabular}{lcccc}
\toprule
\multirow{2}{*}{Method}  & \multicolumn{2}{c}{Endovis2018} & \multicolumn{2}{c}{CholecT50} \\
\cmidrule(lr){2-3} \cmidrule(lr){4-5}
 & F$_1$ & F$_1^{\text{cls}}$ & F$_1$ & F$_1^{\text{cls}}$ \\
\midrule
SFT & 65.7 & 43.2 & 58.7 & \textbf{15.3} \\
RLVR (w/o thinking) & \textbf{67.4} & \textbf{46.3} & \textbf{61.5} & 14.0  \\
\bottomrule
\end{tabular}
  \end{minipage}
  \vspace{-.2cm}
\end{figure*}

In this section, we evaluate our method against SFT-based baselines on the surgical object recognition task.

\vspace{-.3cm}

\paragraph{Baselines.}
We use \texttt{Qwen3-VL-8B-Instruct} as the foundation model and its SFT variant as the primary baseline.  
To analyze each component of our method, we evaluate:  
(1) Cold Start, where pseudo-labeled CoA-style data from unlabeled WebSurg videos are used for unsupervised pre-adaptation;  
(2) Cold Start + SFT, which applies supervised fine-tuning after the cold start; and  
(3) Cold Start + RLVR (CoA), our full method incorporating reinforcement learning with structured reasoning.

\vspace{-.3cm}

\paragraph{Main Results.}
\cref{tab:main} summarizes the performance of all methods on EndoVis2018 and CholecT50.
We report three key remarks derived from these experiments. Qualitative result is provided in the supplementary materials.

\vspace{-.3cm}

\paragraph{Remark 1: CoA outperforms SFT.}
Our \emph{Cold Start + RLVR (CoA)} consistently achieves the highest scores across both benchmarks, substantially surpassing SFT-based baselines.
On EndoVis2018, CoA attains an overall F$_1$ of 83.7 and a per-class macro F$_1^{\text{cls}}$ of 58.0, improving over direct SFT (65.7 / 43.2) by +18.0 and +14.8 points, respectively.
On CholecT50, CoA reaches 64.4 for F$_1$ and 20.2 for F$_1^{\text{cls}}$, outperforming SFT (58.7 / 15.3) by +5.7 and +4.9 points.
Ablation analysis further rules out cold start effects: while \emph{Cold Start + SFT} yields only marginal gains (62.0 vs.\ 65.7 on EndoVis2018), \emph{Cold Start + RLVR (CoA)} consistently delivers the largest improvements across all metrics.
These findings indicate that reinforcement-based optimization (RLVR) provides inherently more stable and robust adaptation than pure supervised fine-tuning, and that the CoA reasoning schema further amplifies this effect.

\vspace{-.3cm}

\paragraph{Remark 2: SFT overfits to dominant classes.}
Across both datasets, SFT achieves relatively high overall F$_1$ but significantly lower F$_1^{\text{cls}}$, reflecting overfitting to frequent categories.
Supervised optimization with cross-entropy tends to favor easy or common classes, leading to class imbalance and weaker generalization to rare instruments.
For instance, on CholecT50 the per-class score drops from 20.7 (Base) to 15.3 (SFT), even as overall F$_1$ rises from 35.3 to 58.7; a similar pattern occurs on EndoVis2018 (43.2 for SFT vs.\ 58.0 for CoA).
The radar plots in \cref{fig:radar} visualize this effect. SFT shows contraction on most challenging classes, while CoA maintains broader and more balanced coverage.

\cref{fig:radar} offers a detailed per-class comparison across both datasets. 
On EndoVis2018 (\cref{fig:radar_endovis}), SFT shows clear collapse on \emph{clamp}, where its F$_1$ scores drop close to zero. 
This issue becomes even more severe on CholecT50 (\cref{fig:radar_cholect50}), with SFT performing poorly on rare tools like \emph{coagulate}, \emph{irrigator}, and \emph{scissors}. These contractions reflect SFT's strong bias toward frequent categories. In contrast, CoA maintains much broader and more balanced coverage in both datasets. Its curves remain extended even on rare and challenging classes, indicating robust generalization and effectively avoiding the severe degradation observed in SFT.


\subsection{Out-of-distribution Surgical Evaluation}

We further evaluate generalization by testing models trained on the in-distribution datasets (EndoVis2018 + CholecT50) on an \emph{unseen} surgical dataset without any additional tuning.
This setting reflects real-world deployment, where surgical procedures, instruments, and imaging conditions differ from the training distribution.

\vspace{-.3cm}

\paragraph{Benchmark.}
GraSP~\citep{ayobi2024pixelwiserecognitionholisticsurgical_grasp} contains 13 robot-assisted radical prostatectomy videos.
For evaluation, we randomly sample 1{,}000 frames to form image–QA pairs following the same transformation pipeline used in the in-distribution experiments.
This benchmark provides a realistic test of model cross-domain generalization to unseen surgical scenes.

\vspace{-.3cm}

\paragraph{Remark 3: CoA generalizes effectively to unseen domains.}
As shown in \cref{tab:ood}, CoA consistently outperforms both the Base and SFT models on GraSP, achieving higher overall F$_1$ and competitive per-class F$_1^{\text{cls}}$.
This demonstrates that CoA’s reinforcement-based optimization and structured reasoning scheme not only enhance in-distribution performance but also transfer effectively to new surgical environments with different procedures and visual contexts.

\vspace{-.3cm}

\paragraph{Remark 4: SFT exhibits cross-domain overfitting.}
In contrast, SFT suffers a significant drop in F$_1^{\text{cls}}$, revealing poor generalization to rare or unseen categories.
The model tends to rely on linguistic and visual priors formed during in-distribution fine-tuning, overemphasizing frequent patterns while neglecting diverse surgical appearances.
This again highlights the limitations of purely supervised adaptation and underscores CoA’s advantage in preserving multimodal robustness through structured reasoning and preference-based learning.

\subsection{Ablation Study}
\label{sec:exp_ablation}

To disentangle the contribution of each component in the proposed CoA framework, we perform a series of controlled ablation experiments.  
Specifically, we analyze how (1) RLVR and (2) the structured reasoning format (CoA vs.\ vanilla CoT)  
each contribute to performance gains in both surgical and general multimodal tasks.

\vspace{-.3cm}

\paragraph{Experimental Variants.}
To ensure fair comparison, all variants are trained under identical hyperparameters, batch sizes, and reward definitions using the same RLVR setup.  
The following configurations are evaluated:

\begin{itemize}
    \item \textbf{SFT:} Conventional SFT using cross-entropy loss.
    \item \textbf{RLVR (w/o thinking):} Directly apply RLVR to unstructured text outputs, without reasoning tags.
    \item \textbf{RLVR + vanilla CoT:} Using the standard \texttt{<thought>}\texttt{</thought>}, \texttt{<answer>}\texttt{</answer>} reasoning format.  We generate 10{,}000 cold start samples with \texttt{Gemini-Flash-2.5} in thinking mode (using a 1{,}024-token reasoning budget) from the same image-question pairs, 
    and subsequently apply the same RLVR optimization procedure as in CoA.
    \item \textbf{RLVR + CoA:} Using the CoA thinking format.
\end{itemize}


\vspace{-.3cm}

\paragraph{Benchmark.} We evaluate all methods on MMBench (\texttt{en-v1.1$_\text{dev}$})~\citep{liu2024mmbenchmultimodalmodelallaround}, which contains 1,164 multiple-choice questions across 20 ability dimensions, and MMStar~\citep{chen2024rightwayevaluatinglarge_mmstar}, a vision-indispensable benchmark with 1,500 human-verified samples designed to assess 6 multimodal capabilities.

\vspace{-.3cm}

\paragraph{Remark 5: RLVR itself surpasses SFT even without structured reasoning.}
\cref{tab:nothinking} compares supervised fine-tuning (SFT) with RLVR applied directly on unstructured responses.  
Even without the CoA reasoning format, RLVR achieves higher overall F$_1$, i.e., 67.4 vs.\ 65.7 on EndoVis2018 and 61.5 vs.\ 58.7 on CholecT50, demonstrating that RLVR provides a more stable and effective adaptation mechanism than SFT.  
However, the per-class F$_1^{\text{cls}}$ slightly drops for CholecT50 (14.0 vs.\ 15.3), suggesting that while RLVR improves robustness, it alone cannot fully address class imbalance or overfitting.  
This underscores the need for additional structure to regularize reasoning and balance generalization across rare categories.

\vspace{-.3cm}

\paragraph{Remark 6: CoA’s structured reasoning enhances both domain and general performance.}
To isolate the effect of structured thinking, we compare RLVR with vanilla CoT and CoA. As shown in \cref{fig:ablation_thinking}, CoA consistently improves both F$_1$ and F$_1^{\text{cls}}$ on EndoVis2018 and CholecT50,  
demonstrating stronger class balance and reduced overfitting to frequent categories. On general benchmarks (MMStar, MMBench), CoA consistently achieves higher accuracy, confirming that the structured reasoning format strengthens surgical adaptation without sacrificing general multimodal competence.

\section{Conclusion}
We study how to adapt VLMs to specialized domains like surgery, where multimodal data are scarce in rich descriptive content and the label space is narrow. We show that RLVR provides a more stable and generalizable alternative to SFT under such challenging conditions. Building on this insight, we introduce CoA, a structured reasoning framework that integrates RLVR with explicit reasoning stages to preserve general multimodal competence while improving domain-specific understanding. Extensive experiments demonstrate that CoA achieves more robust and generalizable adaptation than both SFT and vanilla CoT-based RLVR, providing a promising direction for future research on adapting VLMs to specialized domains.


{
    \small
    \bibliographystyle{ieeenat_fullname}
    \bibliography{main}

@article{wang2025internVL35,
  title={InternVL3.5: Advancing Open-Source Multimodal Models in Versatility, Reasoning, and Efficiency},
  author={Wang, Weiyun and Gao, Zhangwei and Gu, Lixin and Pu, Hengjun and Cui, Long and Wei, Xingguang and others},
  journal={arXiv preprint arXiv:2508.18265},
  year={2025}
}

@article{liu2023visualInstr,
  title={Visual Instruction Tuning},
  author={Liu, Haotian and Li, Chunyuan and Wu, Qingyang and Lee, Yong Jae},
  journal={Advances in Neural Information Processing Systems},
  volume={36},
  year={2024}
}

@article{jin2024surgicalLLAVA,
  title={{S}urgical-{LLaVA}: Toward Surgical Scenario Understanding via Large Language and Vision Models},
  author={Jin, Juseong and Jeong, Chan Woo and others},
  journal={arXiv preprint arXiv:2410.09750},
  year={2024}
}

@article{wang2026endochat,
  title={{E}ndo{C}hat: Grounded multimodal large language model for endoscopic surgery},
  author={Wang, Guankun and Bai, Long and Wang, Junyi and Yuan, Kun and others},
  journal={Medical Image Analysis},
  volume={107},
  pages={103789},
  year={2026},
  note={Epub 2025 Aug 31}
}

@article{schmidgall2024gpvls,
  title={GP-VLS: A general-purpose vision language model for surgery},
  author={Schmidgall, Samuel and Cho, Joseph and Zakka, Cyril and Hiesinger, William},
  journal={arXiv preprint arXiv:2407.19305},
  year={2024}
}

@article{zeng2025surgvlm,
  title={{S}urg{VLM}: A Large Vision-Language Model and Systematic Evaluation Benchmark for Surgical Intelligence},
  author={Zeng, Zhitao and Zhuo, Zhu and Jia, Xiaojun and Zhang, Erli and Wu, Junde and Zhang, Jiaan and Wang, Yuxuan and Low, Chang Han and others},
  journal={arXiv preprint arXiv:2506.02555},
  year={2025}
}

@inproceedings{yang2024llavaMed,
  title={LLaVA-Med: Training a Large Language-and-Vision Assistant for Biomedicine in One Day},
  author={Yang, Jianwei and Naumann, Tristan and Poon, Hoifung and Gao, Jianfeng},
  booktitle={Advances in Neural Information Processing Systems},
  volume={36},
  year={2024}
}

@misc{li2024llavasurgmultimodalsurgicalassistant,
      title={LLaVA-Surg: Towards Multimodal Surgical Assistant via Structured Surgical Video Learning}, 
      author={Jiajie Li and Garrett Skinner and Gene Yang and Brian R Quaranto and Steven D Schwaitzberg and Peter C W Kim and Jinjun Xiong},
      year={2024},
      eprint={2408.07981},
      archivePrefix={arXiv},
      primaryClass={cs.CV},
      url={https://arxiv.org/abs/2408.07981}, 
}

@inproceedings{bai2023surgical,
  title={Surgical-VQLA: Transformer with Gated Vision-Language Embedding for Visual Question Localized-Answering in Robotic Surgery},
  author={Bai, Long and Islam, Mobarakol and Seenivasan, Lalithkumar and Ren, Hongliang},
  booktitle={2023 IEEE International Conference on Robotics and Automation (ICRA)},
  pages={6859--6865},
  year={2023},
  organization={IEEE}
}

@misc{liu2023llava,
      title={Visual Instruction Tuning}, 
      author={Liu, Haotian and Li, Chunyuan and Wu, Qingyang and Lee, Yong Jae},
      publisher={NeurIPS},
      year={2023},
}

@article{Kirkpatrick_2017_catastrophic,
   title={Overcoming catastrophic forgetting in neural networks},
   volume={114},
   ISSN={1091-6490},
   url={http://dx.doi.org/10.1073/pnas.1611835114},
   DOI={10.1073/pnas.1611835114},
   number={13},
   journal={Proceedings of the National Academy of Sciences},
   publisher={Proceedings of the National Academy of Sciences},
   author={Kirkpatrick, James and Pascanu, Razvan and Rabinowitz, Neil and Veness, Joel and Desjardins, Guillaume and Rusu, Andrei A. and Milan, Kieran and Quan, John and Ramalho, Tiago and Grabska-Barwinska, Agnieszka and Hassabis, Demis and Clopath, Claudia and Kumaran, Dharshan and Hadsell, Raia},
   year={2017},
   month=mar, pages={3521–3526} }

@misc{luo2025empiricalstudycatastrophicforgetting,
      title={An Empirical Study of Catastrophic Forgetting in Large Language Models During Continual Fine-tuning}, 
      author={Yun Luo and Zhen Yang and Fandong Meng and Yafu Li and Jie Zhou and Yue Zhang},
      year={2025},
      eprint={2308.08747},
      archivePrefix={arXiv},
      primaryClass={cs.CL},
      url={https://arxiv.org/abs/2308.08747}, 
}

@misc{deepseekai2025deepseekr1incentivizingreasoningcapability,
      title={DeepSeek-R1: Incentivizing Reasoning Capability in LLMs via Reinforcement Learning}, 
      author={DeepSeek-AI and Daya Guo and Dejian Yang and Haowei Zhang and Junxiao Song and Ruoyu Zhang and Runxin Xu and Qihao Zhu and Shirong Ma and Peiyi Wang and Xiao Bi and Xiaokang Zhang and Xingkai Yu and Yu Wu and Z. F. Wu and Zhibin Gou and Zhihong Shao and Zhuoshu Li and Ziyi Gao and Aixin Liu and Bing Xue and Bingxuan Wang and Bochao Wu and Bei Feng and Chengda Lu and Chenggang Zhao and Chengqi Deng and Chenyu Zhang and Chong Ruan and Damai Dai and Deli Chen and Dongjie Ji and Erhang Li and Fangyun Lin and Fucong Dai and Fuli Luo and Guangbo Hao and Guanting Chen and Guowei Li and H. Zhang and Han Bao and Hanwei Xu and Haocheng Wang and Honghui Ding and Huajian Xin and Huazuo Gao and Hui Qu and Hui Li and Jianzhong Guo and Jiashi Li and Jiawei Wang and Jingchang Chen and Jingyang Yuan and Junjie Qiu and Junlong Li and J. L. Cai and Jiaqi Ni and Jian Liang and Jin Chen and Kai Dong and Kai Hu and Kaige Gao and Kang Guan and Kexin Huang and Kuai Yu and Lean Wang and Lecong Zhang and Liang Zhao and Litong Wang and Liyue Zhang and Lei Xu and Leyi Xia and Mingchuan Zhang and Minghua Zhang and Minghui Tang and Meng Li and Miaojun Wang and Mingming Li and Ning Tian and Panpan Huang and Peng Zhang and Qiancheng Wang and Qinyu Chen and Qiushi Du and Ruiqi Ge and Ruisong Zhang and Ruizhe Pan and Runji Wang and R. J. Chen and R. L. Jin and Ruyi Chen and Shanghao Lu and Shangyan Zhou and Shanhuang Chen and Shengfeng Ye and Shiyu Wang and Shuiping Yu and Shunfeng Zhou and Shuting Pan and S. S. Li and Shuang Zhou and Shaoqing Wu and Shengfeng Ye and Tao Yun and Tian Pei and Tianyu Sun and T. Wang and Wangding Zeng and Wanjia Zhao and Wen Liu and Wenfeng Liang and Wenjun Gao and Wenqin Yu and Wentao Zhang and W. L. Xiao and Wei An and Xiaodong Liu and Xiaohan Wang and Xiaokang Chen and Xiaotao Nie and Xin Cheng and Xin Liu and Xin Xie and Xingchao Liu and Xinyu Yang and Xinyuan Li and Xuecheng Su and Xuheng Lin and X. Q. Li and Xiangyue Jin and Xiaojin Shen and Xiaosha Chen and Xiaowen Sun and Xiaoxiang Wang and Xinnan Song and Xinyi Zhou and Xianzu Wang and Xinxia Shan and Y. K. Li and Y. Q. Wang and Y. X. Wei and Yang Zhang and Yanhong Xu and Yao Li and Yao Zhao and Yaofeng Sun and Yaohui Wang and Yi Yu and Yichao Zhang and Yifan Shi and Yiliang Xiong and Ying He and Yishi Piao and Yisong Wang and Yixuan Tan and Yiyang Ma and Yiyuan Liu and Yongqiang Guo and Yuan Ou and Yuduan Wang and Yue Gong and Yuheng Zou and Yujia He and Yunfan Xiong and Yuxiang Luo and Yuxiang You and Yuxuan Liu and Yuyang Zhou and Y. X. Zhu and Yanhong Xu and Yanping Huang and Yaohui Li and Yi Zheng and Yuchen Zhu and Yunxian Ma and Ying Tang and Yukun Zha and Yuting Yan and Z. Z. Ren and Zehui Ren and Zhangli Sha and Zhe Fu and Zhean Xu and Zhenda Xie and Zhengyan Zhang and Zhewen Hao and Zhicheng Ma and Zhigang Yan and Zhiyu Wu and Zihui Gu and Zijia Zhu and Zijun Liu and Zilin Li and Ziwei Xie and Ziyang Song and Zizheng Pan and Zhen Huang and Zhipeng Xu and Zhongyu Zhang and Zhen Zhang},
      year={2025},
      eprint={2501.12948},
      archivePrefix={arXiv},
      primaryClass={cs.CL},
      url={https://arxiv.org/abs/2501.12948}, 
}

@article{shao2024deepseekmath,
  title={Deepseekmath: Pushing the limits of mathematical reasoning in open language models},
  author={Shao, Zhihong and Wang, Peiyi and Zhu, Qihao and Xu, Runxin and Song, Junxiao and Bi, Xiao and Zhang, Haowei and Zhang, Mingchuan and Li, YK and Wu, Yang and others},
  journal={arXiv preprint arXiv:2402.03300},
  year={2024}
}

@article{nwoye2023cholectriplet2022_cholect50,
  title={CholecTriplet2022: Show me a tool and tell me the triplet: an endoscopic vision challenge for surgical action triplet detection.},
  author={Nwoye, Chinedu Innocent and Yu, Tong and Sharma, Saurav and Murali, Aditya and Alapatt, Deepak and Vardazaryan, Armine ... Gonzalez, Cristians and Padoy, Nicolas},
  journal={arXiv preprint arXiv:2204.14746},
  year={2023}
}

@misc{allan20202018roboticscenesegmentation_endovis2018,
      title={2018 Robotic Scene Segmentation Challenge}, 
      author={Max Allan and Satoshi Kondo and Sebastian Bodenstedt and Stefan Leger and Rahim Kadkhodamohammadi and Imanol Luengo and Felix Fuentes and Evangello Flouty and Ahmed Mohammed and Marius Pedersen and Avinash Kori and Varghese Alex and Ganapathy Krishnamurthi and David Rauber and Robert Mendel and Christoph Palm and Sophia Bano and Guinther Saibro and Chi-Sheng Shih and Hsun-An Chiang and Juntang Zhuang and Junlin Yang and Vladimir Iglovikov and Anton Dobrenkii and Madhu Reddiboina and Anubhav Reddy and Xingtong Liu and Cong Gao and Mathias Unberath and Myeonghyeon Kim and Chanho Kim and Chaewon Kim and Hyejin Kim and Gyeongmin Lee and Ihsan Ullah and Miguel Luna and Sang Hyun Park and Mahdi Azizian and Danail Stoyanov and Lena Maier-Hein and Stefanie Speidel},
      year={2020},
      eprint={2001.11190},
      archivePrefix={arXiv},
      primaryClass={cs.CV},
      url={https://arxiv.org/abs/2001.11190}, 
}

@misc{zhao2024swiftascalablelightweightinfrastructure,
      title={SWIFT:A Scalable lightWeight Infrastructure for Fine-Tuning},
      author={Yuze Zhao and Jintao Huang and Jinghan Hu and Xingjun Wang and Yunlin Mao and Daoze Zhang and Zeyinzi Jiang and Zhikai Wu and Baole Ai and Ang Wang and Wenmeng Zhou and Yingda Chen},
      year={2024},
      eprint={2408.05517},
      archivePrefix={arXiv},
      primaryClass={cs.CL},
      url={https://arxiv.org/abs/2408.05517},
}

@inproceedings{kwon2023efficient_vllm,
  title={Efficient Memory Management for Large Language Model Serving with PagedAttention},
  author={Woosuk Kwon and Zhuohan Li and Siyuan Zhuang and Ying Sheng and Lianmin Zheng and Cody Hao Yu and Joseph E. Gonzalez and Hao Zhang and Ion Stoica},
  booktitle={Proceedings of the ACM SIGOPS 29th Symposium on Operating Systems Principles},
  year={2023}
}

@misc{ayobi2024pixelwiserecognitionholisticsurgical_grasp,
      title={Pixel-Wise Recognition for Holistic Surgical Scene Understanding}, 
      author={Nicolás Ayobi and Santiago Rodríguez and Alejandra Pérez and Isabela Hernández and Nicolás Aparicio and Eugénie Dessevres and Sebastián Peña and Jessica Santander and Juan Ignacio Caicedo and Nicolás Fernández and Pablo Arbeláez},
      year={2024},
      eprint={2401.11174},
      archivePrefix={arXiv},
      primaryClass={cs.CV},
      url={https://arxiv.org/abs/2401.11174}, 
}

@misc{zhang2025instructiontuninglargelanguage_sft,
      title={Instruction Tuning for Large Language Models: A Survey}, 
      author={Shengyu Zhang and Linfeng Dong and Xiaoya Li and Sen Zhang and Xiaofei Sun and Shuhe Wang and Jiwei Li and Runyi Hu and Tianwei Zhang and Fei Wu and Guoyin Wang},
      year={2025},
      eprint={2308.10792},
      archivePrefix={arXiv},
      primaryClass={cs.CL},
      url={https://arxiv.org/abs/2308.10792}, 
}

@misc{shumailov2024curserecursiontraininggenerated_collapse,
      title={The Curse of Recursion: Training on Generated Data Makes Models Forget}, 
      author={Ilia Shumailov and Zakhar Shumaylov and Yiren Zhao and Yarin Gal and Nicolas Papernot and Ross Anderson},
      year={2024},
      eprint={2305.17493},
      archivePrefix={arXiv},
      primaryClass={cs.LG},
      url={https://arxiv.org/abs/2305.17493}, 
}

@misc{comanici2025gemini25pushingfrontier,
      title={Gemini 2.5: Pushing the Frontier with Advanced Reasoning, Multimodality, Long Context, and Next Generation Agentic Capabilities}, 
      author={DeepMind},
      year={2025},
      eprint={2507.06261},
      archivePrefix={arXiv},
      primaryClass={cs.CL},
      url={https://arxiv.org/abs/2507.06261}, 
}

@article{Qwen2-VL,
  title={Qwen2-VL: Enhancing Vision-Language Model's Perception of the World at Any Resolution},
  author={Wang, Peng and Bai, Shuai and Tan, Sinan and Wang, Shijie and Fan, Zhihao and Bai, Jinze and Chen, Keqin and Liu, Xuejing and Wang, Jialin and Ge, Wenbin and Fan, Yang and Dang, Kai and Du, Mengfei and Ren, Xuancheng and Men, Rui and Liu, Dayiheng and Zhou, Chang and Zhou, Jingren and Lin, Junyang},
  journal={arXiv preprint arXiv:2409.12191},
  year={2024}
}

@article{Qwen-VL,
  title={Qwen-VL: A Versatile Vision-Language Model for Understanding, Localization, Text Reading, and Beyond},
  author={Bai, Jinze and Bai, Shuai and Yang, Shusheng and Wang, Shijie and Tan, Sinan and Wang, Peng and Lin, Junyang and Zhou, Chang and Zhou, Jingren},
  journal={arXiv preprint arXiv:2308.12966},
  year={2023}
}

@inproceedings{chen2024internvl,
  title={Internvl: Scaling up vision foundation models and aligning for generic visual-linguistic tasks},
  author={Chen, Zhe and Wu, Jiannan and Wang, Wenhai and Su, Weijie and Chen, Guo and Xing, Sen and Zhong, Muyan and Zhang, Qinglong and Zhu, Xizhou and Lu, Lewei and others},
  booktitle={Proceedings of the IEEE/CVF Conference on Computer Vision and Pattern Recognition},
  pages={24185--24198},
  year={2024}
}

@misc{kimiteam2025kimivltechnicalreport,
      title={Kimi-VL Technical Report}, 
      author={Kimi Team and Angang Du and Bohong Yin and Bowei Xing and Bowen Qu and Bowen Wang and Cheng Chen and Chenlin Zhang and Chenzhuang Du and Chu Wei and Congcong Wang and Dehao Zhang and Dikang Du and Dongliang Wang and Enming Yuan and Enzhe Lu and Fang Li and Flood Sung and Guangda Wei and Guokun Lai and Han Zhu and Hao Ding and Hao Hu and Hao Yang and Hao Zhang and Haoning Wu and Haotian Yao and Haoyu Lu and Heng Wang and Hongcheng Gao and Huabin Zheng and Jiaming Li and Jianlin Su and Jianzhou Wang and Jiaqi Deng and Jiezhong Qiu and Jin Xie and Jinhong Wang and Jingyuan Liu and Junjie Yan and Kun Ouyang and Liang Chen and Lin Sui and Longhui Yu and Mengfan Dong and Mengnan Dong and Nuo Xu and Pengyu Cheng and Qizheng Gu and Runjie Zhou and Shaowei Liu and Sihan Cao and Tao Yu and Tianhui Song and Tongtong Bai and Wei Song and Weiran He and Weixiao Huang and Weixin Xu and Xiaokun Yuan and Xingcheng Yao and Xingzhe Wu and Xinhao Li and Xinxing Zu and Xinyu Zhou and Xinyuan Wang and Y. Charles and Yan Zhong and Yang Li and Yangyang Hu and Yanru Chen and Yejie Wang and Yibo Liu and Yibo Miao and Yidao Qin and Yimin Chen and Yiping Bao and Yiqin Wang and Yongsheng Kang and Yuanxin Liu and Yuhao Dong and Yulun Du and Yuxin Wu and Yuzhi Wang and Yuzi Yan and Zaida Zhou and Zhaowei Li and Zhejun Jiang and Zheng Zhang and Zhilin Yang and Zhiqi Huang and Zihao Huang and Zijia Zhao and Ziwei Chen and Zongyu Lin},
      year={2025},
      eprint={2504.07491},
      archivePrefix={arXiv},
      primaryClass={cs.CV},
      url={https://arxiv.org/abs/2504.07491}, 
}

@misc{wu2024gpt4visionhumanalignedevaluatortextto3d,
      title={GPT-4V(ision) is a Human-Aligned Evaluator for Text-to-3D Generation}, 
      author={Tong Wu and Guandao Yang and Zhibing Li and Kai Zhang and Ziwei Liu and Leonidas Guibas and Dahua Lin and Gordon Wetzstein},
      year={2024},
      eprint={2401.04092},
      archivePrefix={arXiv},
      primaryClass={cs.CV},
      url={https://arxiv.org/abs/2401.04092}, 
}

@misc{ouyang2022traininglanguagemodelsfollow_rlhf,
      title={Training language models to follow instructions with human feedback}, 
      author={Long Ouyang and Jeff Wu and Xu Jiang and Diogo Almeida and Carroll L. Wainwright and Pamela Mishkin and Chong Zhang and Sandhini Agarwal and Katarina Slama and Alex Ray and John Schulman and Jacob Hilton and Fraser Kelton and Luke Miller and Maddie Simens and Amanda Askell and Peter Welinder and Paul Christiano and Jan Leike and Ryan Lowe},
      year={2022},
      eprint={2203.02155},
      archivePrefix={arXiv},
      primaryClass={cs.CL},
      url={https://arxiv.org/abs/2203.02155}, 
}

@misc{wang2024rlvlmfreinforcementlearningvision_rl_vlm,
      title={RL-VLM-F: Reinforcement Learning from Vision Language Foundation Model Feedback}, 
      author={Yufei Wang and Zhanyi Sun and Jesse Zhang and Zhou Xian and Erdem Biyik and David Held and Zackory Erickson},
      year={2024},
      eprint={2402.03681},
      archivePrefix={arXiv},
      primaryClass={cs.RO},
      url={https://arxiv.org/abs/2402.03681}, 
}

@misc{chen2024visionlanguagemodelsprovidepromptable,
      title={Vision-Language Models Provide Promptable Representations for Reinforcement Learning}, 
      author={William Chen and Oier Mees and Aviral Kumar and Sergey Levine},
      year={2024},
      eprint={2402.02651},
      archivePrefix={arXiv},
      primaryClass={cs.LG},
      url={https://arxiv.org/abs/2402.02651}, 
}

@misc{wei2023chainofthoughtpromptingelicitsreasoning,
      title={Chain-of-Thought Prompting Elicits Reasoning in Large Language Models}, 
      author={Jason Wei and Xuezhi Wang and Dale Schuurmans and Maarten Bosma and Brian Ichter and Fei Xia and Ed Chi and Quoc Le and Denny Zhou},
      year={2023},
      eprint={2201.11903},
      archivePrefix={arXiv},
      primaryClass={cs.CL},
      url={https://arxiv.org/abs/2201.11903}, 
}

@misc{chen2025g1bootstrappingperceptionreasoning,
      title={G1: Bootstrapping Perception and Reasoning Abilities of Vision-Language Model via Reinforcement Learning}, 
      author={Liang Chen and Hongcheng Gao and Tianyu Liu and Zhiqi Huang and Flood Sung and Xinyu Zhou and Yuxin Wu and Baobao Chang},
      year={2025},
      eprint={2505.13426},
      archivePrefix={arXiv},
      primaryClass={cs.CV},
      url={https://arxiv.org/abs/2505.13426}, 
}

@misc{liu2024mmbenchmultimodalmodelallaround,
      title={MMBench: Is Your Multi-modal Model an All-around Player?}, 
      author={Yuan Liu and Haodong Duan and Yuanhan Zhang and Bo Li and Songyang Zhang and Wangbo Zhao and Yike Yuan and Jiaqi Wang and Conghui He and Ziwei Liu and Kai Chen and Dahua Lin},
      year={2024},
      eprint={2307.06281},
      archivePrefix={arXiv},
      primaryClass={cs.CV},
      url={https://arxiv.org/abs/2307.06281}, 
}

@misc{chen2024rightwayevaluatinglarge_mmstar,
      title={Are We on the Right Way for Evaluating Large Vision-Language Models?}, 
      author={Lin Chen and Jinsong Li and Xiaoyi Dong and Pan Zhang and Yuhang Zang and Zehui Chen and Haodong Duan and Jiaqi Wang and Yu Qiao and Dahua Lin and Feng Zhao},
      year={2024},
      eprint={2403.20330},
      archivePrefix={arXiv},
      primaryClass={cs.CV},
      url={https://arxiv.org/abs/2403.20330}, 
}

@misc{twinanda2016endonetdeeparchitecturerecognition,
      title={EndoNet: A Deep Architecture for Recognition Tasks on Laparoscopic Videos}, 
      author={Andru P. Twinanda and Sherif Shehata and Didier Mutter and Jacques Marescaux and Michel de Mathelin and Nicolas Padoy},
      year={2016},
      eprint={1602.03012},
      archivePrefix={arXiv},
      primaryClass={cs.CV},
      url={https://arxiv.org/abs/1602.03012}, 
}

@article{stauder2016tum,
  title={The TUM LapChole dataset for the M2CAI 2016 workflow challenge},
  author={Stauder, Ralf and Ostler, Daniel and Kranzfelder, Michael and Koller, Sebastian and Feu{\ss}ner, Hubertus and Navab, Nassir},
  journal={arXiv preprint arXiv:1610.09278},
  year={2016}
}

@article{maier2021heidelberg,
  title={Heidelberg colorectal data set for surgical data science in the sensor operating room},
  author={Maier-Hein, Lena and Wagner, Martin and Ross, Tobias and Reinke, Annika and Bodenstedt, Sebastian and Full, Peter M and Hempe, Hellena and Mindroc-Filimon, Diana and Scholz, Patrick and Tran, Thuy Nuong and others},
  journal={Scientific data},
  volume={8},
  number={1},
  pages={101},
  year={2021},
  publisher={Nature Publishing Group UK London}
}

@article{huaulme2021micro,
  title={Micro-surgical anastomose workflow recognition challenge report},
  author={Huaulm{\'e}, Arnaud and Sarikaya, Duygu and Le Mut, K{\'e}vin and Despinoy, Fabien and Long, Yonghao and Dou, Qi and Chng, Chin-Boon and Lin, Wenjun and Kondo, Satoshi and Bravo-S{\'a}nchez, Laura and others},
  journal={Computer Methods and Programs in Biomedicine},
  volume={212},
  pages={106452},
  year={2021},
  publisher={Elsevier}
}

@inproceedings{wang2022autolaparo,
  title={Autolaparo: A new dataset of integrated multi-tasks for image-guided surgical automation in laparoscopic hysterectomy},
  author={Wang, Ziyi and Lu, Bo and Long, Yonghao and Zhong, Fangxun and Cheung, Tak-Hong and Dou, Qi and Liu, Yunhui},
  booktitle={International Conference on Medical Image Computing and Computer-Assisted Intervention},
  pages={486--496},
  year={2022},
  organization={Springer}
}

@article{wagner2023comparative,
  title={Comparative validation of machine learning algorithms for surgical workflow and skill analysis with the heichole benchmark},
  author={Wagner, Martin and M{\"u}ller-Stich, Beat-Peter and Kisilenko, Anna and Tran, Duc and Heger, Patrick and M{\"u}ndermann, Lars and Lubotsky, David M and M{\"u}ller, Benjamin and Davitashvili, Tornike and Capek, Manuela and others},
  journal={Medical image analysis},
  volume={86},
  pages={102770},
  year={2023},
  publisher={Elsevier}
}

@article{van2022gesture,
  title={Gesture recognition in robotic surgery with multimodal attention},
  author={Van Amsterdam, Beatrice and Funke, Isabel and Edwards, Eddie and Speidel, Stefanie and Collins, Justin and Sridhar, Ashwin and Kelly, John and Clarkson, Matthew J and Stoyanov, Danail},
  journal={IEEE transactions on medical imaging},
  volume={41},
  number={7},
  pages={1677--1687},
  year={2022},
  publisher={IEEE}
}

@article{hao2025surgery_r1,
  title={Surgery-R1: Advancing Surgical-VQLA with Reasoning Multimodal Large Language Model via Reinforcement Learning},
  author={Hao, Pengfei and Li, Shuaibo and Wang, Hongqiu and Kou, Zhizhuo and Zhang, Junhang and Yang, Guang and Zhu, Lei},
  journal={arXiv preprint arXiv:2506.19469},
  year={2025}
}

@InProceedings{pmlr-v267-chu25c,
  title = 	 {{SFT} Memorizes, {RL} Generalizes: A Comparative Study of Foundation Model Post-training},
  author =       {Chu, Tianzhe and Zhai, Yuexiang and Yang, Jihan and Tong, Shengbang and Xie, Saining and Schuurmans, Dale and Le, Quoc V and Levine, Sergey and Ma, Yi},
  booktitle = 	 {Proceedings of the 42nd International Conference on Machine Learning},
  pages = 	 {10818--10838},
  year = 	 {2025},
  volume = 	 {267},
  series = 	 {Proceedings of Machine Learning Research},
  month = 	 {13--19 Jul},
  publisher =    {PMLR},
  url = 	 {https://proceedings.mlr.press/v267/chu25c.html}
}

@inproceedings{
yue2025does,
title={Does Reinforcement Learning Really Incentivize Reasoning Capacity in {LLM}s Beyond the Base Model?},
author={Yang Yue and Zhiqi Chen and Rui Lu and Andrew Zhao and Zhaokai Wang and Yang Yue and Shiji Song and Gao Huang},
booktitle={The Thirty-ninth Annual Conference on Neural Information Processing Systems},
year={2025},
url={https://openreview.net/forum?id=4OsgYD7em5}
}

@article{qwen3,
    title={Qwen3 Technical Report}, 
    author={An Yang and Anfeng Li and Baosong Yang and Beichen Zhang and Binyuan Hui and Bo Zheng and Bowen Yu and Chang Gao and Chengen Huang and Chenxu Lv and Chujie Zheng and Dayiheng Liu and Fan Zhou and Fei Huang and Feng Hu and Hao Ge and Haoran Wei and Huan Lin and Jialong Tang and Jian Yang and Jianhong Tu and Jianwei Zhang and Jianxin Yang and Jiaxi Yang and Jing Zhou and Jingren Zhou and Junyang Lin and Kai Dang and Keqin Bao and Kexin Yang and Le Yu and Lianghao Deng and Mei Li and Mingfeng Xue and Mingze Li and Pei Zhang and Peng Wang and Qin Zhu and Rui Men and Ruize Gao and Shixuan Liu and Shuang Luo and Tianhao Li and Tianyi Tang and Wenbiao Yin and Xingzhang Ren and Xinyu Wang and Xinyu Zhang and Xuancheng Ren and Yang Fan and Yang Su and Yichang Zhang and Yinger Zhang and Yu Wan and Yuqiong Liu and Zekun Wang and Zeyu Cui and Zhenru Zhang and Zhipeng Zhou and Zihan Qiu},
    journal = {arXiv preprint arXiv:2505.09388},
    year={2025}
}
}



\end{document}